\def\year{2018}\relax
\documentclass[letterpaper]{article} 
\usepackage{aaai18}  
\usepackage{times}  
\usepackage{helvet}  
\usepackage{courier}  
\usepackage{url}  
\usepackage{graphicx}  
\usepackage{multirow}
\usepackage{amsfonts}
\usepackage{caption}
\usepackage{subcaption}
\usepackage{color}
\usepackage[ruled,vlined]{algorithm2e}

\newcommand{\ndcg}{{\textsc{NDCG@30}}}
\newcommand{\pre}{{\textsc{P@5}}}
\newcommand{\rec}{{\textsc{R@30}}}

\newcommand{\lightfm}{\textsc{LightFM}}

\newcommand{\pop}{\textsc{POP}}

\newcommand{\awarp}{\textsc{WARP}}

\newcommand{\bbpr}{b-BPR}
\newcommand{\bpr}{BPR}

\newcommand{\xing}{\textsf{XING}}
\newcommand{\yelp}{\textsf{Yelp}}

\newcommand{\movietwenty}{\textsf{ML-20m}}

\begin{document}
%
\title{A Batch Learning Framework for Scalable Personalized Ranking}
\author{Kuan Liu, Prem Natarajan\\
Information Sciences Institute \& Computer Science Department\\
University of Southern California\\
}
\maketitle
\begin{abstract}
In designing personalized ranking algorithms, it is desirable to encourage a high precision at the top of the ranked list. Existing methods either seek a smooth convex surrogate for a non-smooth ranking metric or directly modify updating procedures to encourage top accuracy. In this work we point out that these methods do not scale well to a large-scale setting, and this is partly due to the inaccurate pointwise or pairwise rank estimation. We propose a new framework for personalized ranking. It uses batch-based rank estimators and smooth rank-sensitive loss functions. This new batch learning framework leads to more stable and accurate rank approximations compared to previous work. Moreover, it enables explicit use of parallel computation to speed up training. We conduct empirical evaluation on three item recommendation tasks. Our method shows consistent accuracy improvements over state-of-the-art methods. Additionally, we observe time efficiency advantages when data scale increases.


\end{abstract}

\section{Introduction}

The task of personalized ranking is to provide each user a ranked list of items that he might prefer. It has received much attention in academic research~\cite{hu2008collaborative,rendle2009bpr,he2016fast}, and algorithms developed are applied in various applications in e-commerce~\cite{linden2003amazon}, social networks~\cite{chen2009make}, location~\cite{liu2014exploiting}, etc. However, personalized ranking remains a very challenging task: 1) The learning objectives of the ranking models are hard to directly optimize. For example, the quality of the model output is commonly evaluated by ranking measures such as NDCG, MAP, MRR, which are position-dependent (or rank-dependent) and non-smooth. It makes gradient-based optimization infeasible and also computationally expensive. 2) The size of an item set that a ranking task uses can be very large. It is not uncommon to see an online recommender system with millions of items. As a consequence, it increases the difficulty of capturing user preferences over the entire set of items. It also makes it harder to compute or estimate the rank of a particular item.

Traditional approaches model user preferences with \textit{rank-independent} algorithms. Pairwise algorithms convert the learning task into many binary classification problems and optimize the \textit{average classification accuracy}. For example, BPR~\cite{rendle2009bpr,rendle2014improving} maximizes the probability of correct prediction of each pairwise item comparison. MMMF~\cite{srebro2005maximum}~minimizes a margin loss for each pair in a matrix factorization setting. Listwise algorithms such as those recently explored by~\cite{hidasi2015session,covington2016deep} treat the problem as a multi-class classification and use cross-entropy as the loss function.

Despite the simplicity and wide application, these rank-independent methods are not satisfactory because the quality of results from a ranking system is highly position-dependent. A high accuracy at the top of a list is more important than that at a low position on the list. However, the average accuracy targeted by the pairwise algorithms discussed above places equal weights at all the positions. This mismatch therefore leads to under-exploitation in the prediction accuracy at the top part. Listwise algorithms, on the other hand, do make an attempt to push items to the top by a classification scheme. However, its classification criterion also does not match well with ranking.

Position-dependent approaches are explored to address the above limitations. One critical question is how to approximate item ranks to perform rank-dependent training. TFMAP~\cite{shi2012tfmap}~and~ClifMF~\cite{shi2012climf}~approximate an item rank purely based on the model score of this item, i.e., a \textbf{pointwise estimate}. Particularly, they model the reciprocal rank of an item with a sigmoid transformed from the score returned by the model. TFMAP then optimizes a smoothed modification of MAP, while ClifMF optimizes MRR. This pointwise estimation is simple but is only loosely connected to the true ranks. The estimation becomes even more unreliable as the itemset size increases.

An arguably more direct approach is to optimize the ranks of relevant items returned by the model to encourage top accuracy. It requires the computation or estimation of item ranks and modification of the updating strategy. This idea is explored in traditional learning to rank methods LambdaNet~\cite{burges2005learning}, LambdaRank~\cite{burges2007learning}, etc., where the learning rate is adjusted based on item ranks. In personalized ranking, WARP~\cite{weston2010large}~proposes to use a sampling procedure to estimate item ranks. It repeatedly samples negative items until it finds one that has a higher score. Then the number of sampling trials is used to estimate item ranks. This stochastic \textbf{pairwise estimation} is intuitive. WARP is also found to be more competitive than BPR~\cite{hong2013co}. A more recent matrix factorization model LambdaFM~\cite{yuan2016lambdafm}~adopts the same rank estimation. However, this pairwise rank estimator becomes very noisy and unstable as the itemset size increases as we will demonstrate. It takes a large number of sampling iterations for each estimation. Moreover, its intrinsic online learning fashion prevents full utilization of available parallel computation (e.g., GPUs), making it hard to train large or flexible models which rely on the parallel computation. 

The limitations of these approaches largely come from the stochastic pairwise estimation component. As a comparison, training with batch or mini-batch together with parallel computation has recently offered opportunities to tackle scalability challenges. \cite{hidasi2015session}~uses a sampled classification setting to train a RNNs-based recommender system .~\cite{covington2016deep}~deploy a two-stage classification system in YouTube video recommendation. Parallel computation (e.g., GPUs) are extensively used to accelerate training and support flexible models.

In this work we propose a novel framework to do personalized ranking. Our aim is to have better rank approximations and advocate the top accuracy in large-scale settings. The contributions of this work are:

\begin{itemize}
\item We propose rank estimations that are based on batch computation. They are shown to be more accurate and stable in approximating the true item rank.
\item We propose smooth loss functions that are ``rank-sensitive.'' This advocates top accuracy by optimizing the loss function values. Being differentiable, the functions are easily implemented and integrated with back-propagation updates.
\item Based on batch computation, the algorithms explicitly utilize parallel computation to speed up training. They lend themselves to flexible models which rely on more extensive computation.
\item Our experiments on three item recommendation tasks show consistent accuracy improvements over state-of-the-art algorithms. They also show time efficiency advantages when data scale increases.
\end{itemize}

The remainder of the paper is organized as follows: We first introduce notations and preliminary methods. We next detail our new methods, followed by discussions on related work. Experimental results are then presented. We conclude with discussions and future work.

\section{Notations}
\label{sec:notation}

In this paper we will use the letter $x$ for users and the letter $y$ for items. We use unbolded letters to denote single elements of users or items and bolded letters to denote a set of items. Particularly, a single user and a single item are, respectively, denoted by $x$ and $y$, \textbf{Y} denotes the entire item set. $\textbf{y}_x$ denotes the positive items of user $x$---that is, the subset of items that are interacted by user $x$. $\bar{\textbf{y}}_x\equiv\textbf{Y} \setminus \textbf{y}_x$ is the irrelevant item set of user $x$. We omit subscript x when there is no ambiguity. $\textbf{S}=\{(x,y)\}$ is the set of observations. The indicator function is denoted by \textbf{I}. f denotes a model (or model parameters).  $f_y(x)$ denotes the model score for user $x$ and item $y$. For example, in a matrix factorization model, $f_y(x)$ is computed by the inner product of latent factors of $x$ and $y$. Given a model $f$, user $x$, and its positive items $\textbf{y}$, the rank of an item $y$ is defined as 
\begin{equation}
\label{eq:rank}
r_y\equiv rank_y(f,x,\textbf{y}) =\sum_{y'\in\bar{\textbf{y}}} \textbf{I}[f_y(x) \le f_{y'}(x)],
\end{equation}
where we use the same definition as in \cite{usunier2009ranking} and ignore the order within positive items. The indicator function is sometimes approximated by a continuous margin loss $|1-f_y(x) + f_{y'}(x)|_+$ where $|t|_+ \equiv max(t,0)$.

\section{Position-dependent personalized ranking}


Position-dependent algorithms take the ranks of predicted items into account in the model training. A practical challenge is how to estimate item ranks efficiently. As seen in (\ref{eq:rank}), the ranks depend on the model parameters and are dynamically changing. The definition is non-smooth in model parameters due to the indicator function. The computation of ranks involves comparisons with all the irrelevant items, which can be costly. 

Existing position-dependent algorithms address the challenge by different rank approximation methods. They can be categorized into pointwise and pairwise approximations. We describe their approaches in the following.

\subsection{Pointwise rank approximation}
Item ranks are approximated in TFMAP~\cite{shi2012tfmap}~and~ClifMF~\cite{shi2012climf} via an Pointwise approach. Particularly, 

\begin{equation}
\label{eq:element}
r_y \approx rank^{point}_y(f,x) = 1 / \sigma (f_y(x)),
\end{equation}
where $\sigma(z) = 1 / (1 + e^{-z}), \forall z\in \mathbb{R}$. The rank $r_y$ is then plugged into an evaluation metric MAP (as in TFMAP) and MRR (as in ClifMF) to make the objective smooth. The algorithms then use gradient-based methods for optimization.

In (\ref{eq:element}), $rank^{point}_y$ is close to 1 when model score $f_y(x)$ is high and becomes large when $f_y(x)$ is low. This connection between model scores and item ranks is intuitive, and implicitly encourages a good accuracy at the top. However, $rank^{point}_y$ is very loosely connected to rank definition in (\ref{eq:rank}). In practice, it does not capture the non-smooth characteristics of ranks. For example, small differences in model scores can lead to dramatic rank differences when the item set is large.

\subsection{Pairwise rank approximation}
An alternative approach used by WARP~\cite{weston2010large}, LambdaMF~\cite{yuan2016lambdafm}~estimates item ranks based on comparisons between a pair of items. The critical component is an \textbf{iterative sampling approximation procedure}: given a user x and a positive item y, keep sampling a negative item $y'\in\bar{\textbf{y}}$ uniformly with replacement until the condition $1+f_{y'}(x)<f_y(x)$ is violated. With the sampling procedure it estimates item ranks by

\begin{equation}
\label{eq:warp3}
r_y \approx rank^{pair}_y(f,x,\textbf{y}) = \lfloor \frac{|\bar{\textbf{y}}|-1}{N} \rfloor
\end{equation}
where N is the number of sampling trials to find the first violating example and $\lfloor z \rfloor$ takes the maximum integer that is no greater than $z$.

The intuition behind this estimation is that the number of trials of sampling follows a geometric distribution. Suppose the item's true rank is $r_y$, the probability of having a violating item is $p = r_y / (|\bar{\textbf{y}}| - 1)$. The expected number of trials is $\mathbb{E}(N) = 1/p=(|\bar{\textbf{y}}| - 1)/r_y$.

To promote top accuracy, the estimated item ranks are used to modify updating procedures. For example, in WARP, they are plugged in a loss function defined as, 

\begin{equation}
\label{eq:owc}
L^{owa}(x,y) = \Phi^{owa}[r_y] = \sum_{j=1}^{r_y} \alpha_j \quad \alpha_1\ge \alpha_2\ge..\ge 0
\end{equation}
where $\alpha_j$ $j=1,2,..$ are predefined non-increasing scalars. The function $\Phi^{owa}$ is derived from ordered weighted average (OWA) of classification loss ~\cite{yager1988ordered} in ~\cite{usunier2009ranking}. It defines a penalty incurred by placing $r_y$ irrelevant elements before a relevant one. Choosing equal $\alpha_j$ means optimizing the mean rank, while choosing $\alpha_{j>1}=0, \alpha_1=1$ means optimizing the proportion of top-ranked correct items. With strictly decreasing $\alpha$s, it optimizes the top part of a ranked list.

\section{Approach}
\label{sec:method}
We begin by pointing out several limitations of approaches based on pairwise rank estimations.

First, the rank estimator in (\ref{eq:warp3}) is not only biased but also has a large variance. Expectation of the estimator in (\ref{eq:warp3}) for $p$ of a geometric distribution is approximately $p (1+ \sum_{k=2}^N \frac{1}{k} (1-p)^{k-1}) > p$. In a ranking example where $p = r / N$ (N population size), it overestimates the rank seriously when r is small. Moreover, we will demonstrate later that the estimator has high estimation variances. We believe this poor estimation may lead to training inefficiency. 
Second, it can take a large number of sampling iterations before finding a violating item in each iteration. This is especially the case after the beginning stage of training. This results in low frequency of model updating. In practice, prolonged training time is observed.
Finally, the \textit{intrinsic sequential learning fashion} based on pairwise estimation prevents full utilization of available parallel computation (e.g., GPUs). This makes it hard to train large or flexible models which highly rely on the parallel computation.


We address the limitations by combining the ideas of batch computation and rank-dependent training loss. Particularly, we propose batch rank approximations and generalize (\ref{eq:owc}) to smooth rank-sensitive loss functions. The resulting algorithm gives more accurate rank approximations and allows back-propagation updates.

\subsection{Batch rank approximations}

In order to have a stable and accurate rank approximation that leads to efficient algorithms. We exploit the idea of batch training that is recently actively explored or revisited such as in model design~\cite{covington2016deep}.

\begin{figure}
\centering
\begin{subfigure}{.25\textwidth}
  \centering
  \includegraphics[width=.99\linewidth]{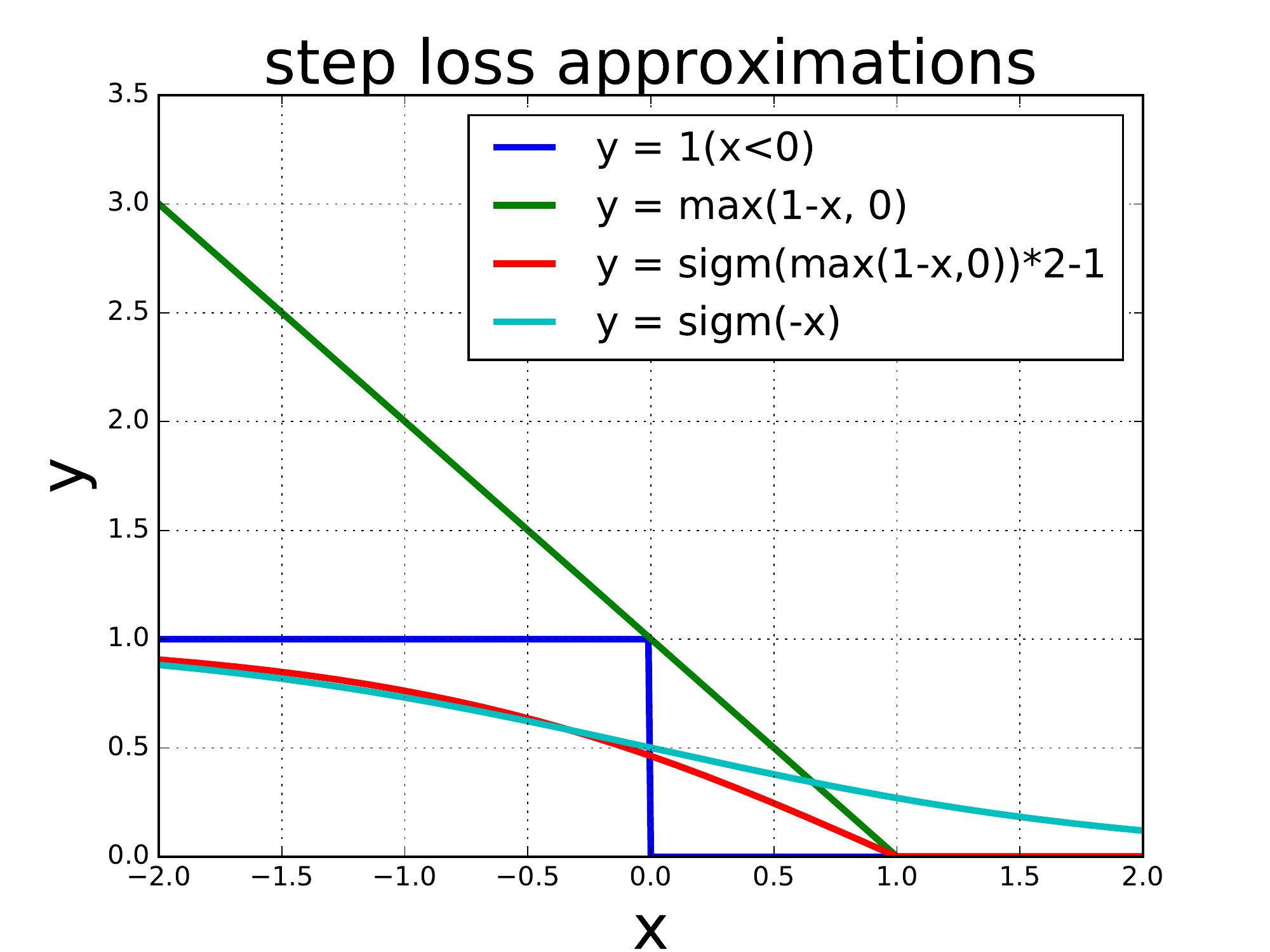}
  \caption{Indicator approximations.}
  \label{fig:rapp}
\end{subfigure}%
\begin{subfigure}{.25\textwidth}
  \centering
  \includegraphics[width=.99\linewidth]{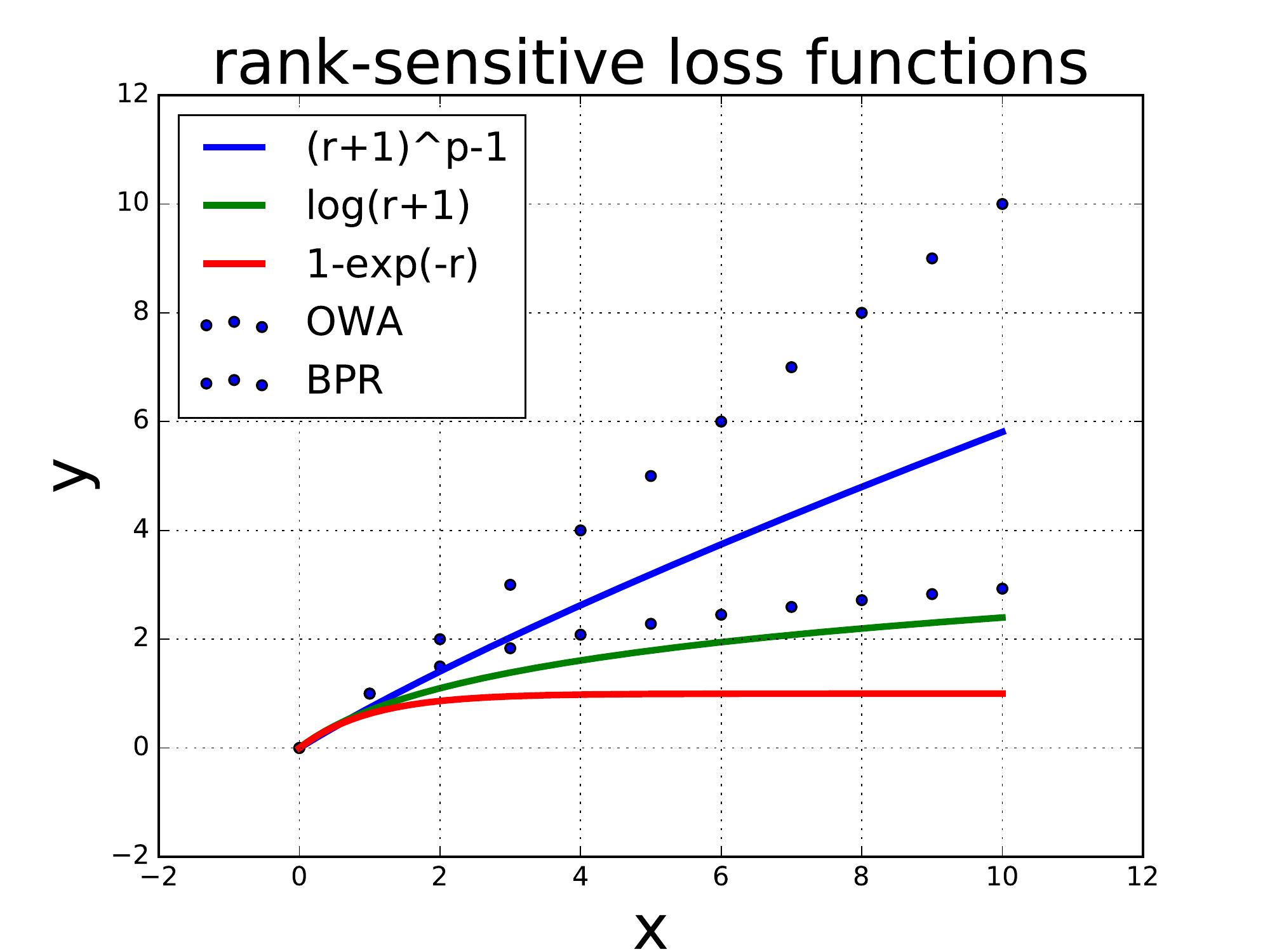}
  \caption{Rank-sensitive loss functions.}
  \label{fig:rsl}
\end{subfigure}
\caption{Illustrations of rank approximations and smooth rank-sensitive loss functions. \ref{fig:rapp} shows different approximations of indicator functions. \ref{fig:rsl} shows smooth loss functions used to generalize the loss (\ref{eq:owc}).}
\label{fig:approach}
\end{figure}

To begin, we define \textit{margin rank} (mr) as the following:

\begin{equation}
\label{eq:mr}
rank^{mr}_y(f,x,\textbf{y}) =\sum_{y'\in \bar{\textbf{y}} } |1-f_y(x) + f_{y'}(x)|_+, 
\end{equation}
where the margin loss is used to replace the indicator function in (\ref{eq:rank}), and the target item $y$ is compared to a batch of negative items $\bar{\textbf{y}}$.
As illustrated in Figure \ref{fig:rapp}, the margin loss (green curve) is a smooth convex upper bound of the original step loss function (or indicator function). \textit{Margin rank} sums up the margin errors and characterizes the overall violation of irrelevant items.

\textit{Margin rank} could be sensitive to ``bad'' items that have significantly higher scores than the target item. As seen in Eq. (\ref{eq:mr}) or Figure \ref{fig:rapp}, such a bad item contributes much more than one in rank estimation. To suppress that effect, we add a sigmoid transformation, i.e.,
\begin{equation}
\label{eq:smr}
rank^{smr}_y(f,x,\textbf{y}) =\sum_{y'\in \bar{\textbf{y}} } 2 * \sigma(|1-f_y(x) + f_{y'}(x)|_+) + 1, 
\end{equation}
where $\sigma(z) = 1 / (1 + e^{-z}), \forall z\in \mathbb{R}$. We call this \textit{suppressed margin rank} (smr). Additionally, we study a smoother version without margin formulation, i.e., $rank^{sr}_y(f,x,\textbf{y}) =\sum_{y'\in \bar{\textbf{y}} } \sigma( f_{y'}(x) - f_{y}(x))$. Therefore, our rank approximations can be written as

\begin{equation}
\label{eq:batch}
rank^{batch}_y(f,x,\textbf{y}) =\sum_{y'\in \bar{\textbf{y}} } \tilde{r}(x,y,y'),
\end{equation}
where $\tilde{r}(x, y, y')$ takes one of the following three forms:
\begin{itemize}
\item $\tilde{r}(x, y, y') = | 1 - f_y(x) + f_{y'}(x)|_+$
\item $\tilde{r}(x, y, y') = 2 * \sigma(| 1 - f_y(x) + f_{y'}(x)|_+)$  - 1
\item $\tilde{r}(x, y, y') = \sigma(f_{y'}(x) - f_y(x))$
\end{itemize}

Note that the rank approximations in (\ref{eq:batch}) are computed in a batch manner: model scores between a user and a batch of items are first computed in parallel; $\tilde{r}$ are then computed accordingly and summed up. This batch computation allows model parameters to be updated more frequently. The parallel computation can speed up model training. 

The full batch treatment may become infeasible or inefficient when the item set gets overly large. A mini-batch approach is used for that case. Instead of computing the rank based on the full set $\textbf{Y}$ as in (\ref{eq:batch}), the mini-batch version algorithm samples $\textbf{Z}$, a subset of $\textbf{Y}$ randomly (without replacement) and computes

\begin{equation}
\label{eq:mrmini}
rank^{mb}_y(f,x,\textbf{y}) = \frac{|\textbf{Y}|}{|\textbf{Z}|}\sum_{y'\in \textbf{Z}} \tilde{r}(x,y,y') \textbf{I} (y'\in\bar{\textbf{y}}).
\end{equation}

\begin{figure}
\centering
\includegraphics[width=0.89\columnwidth]{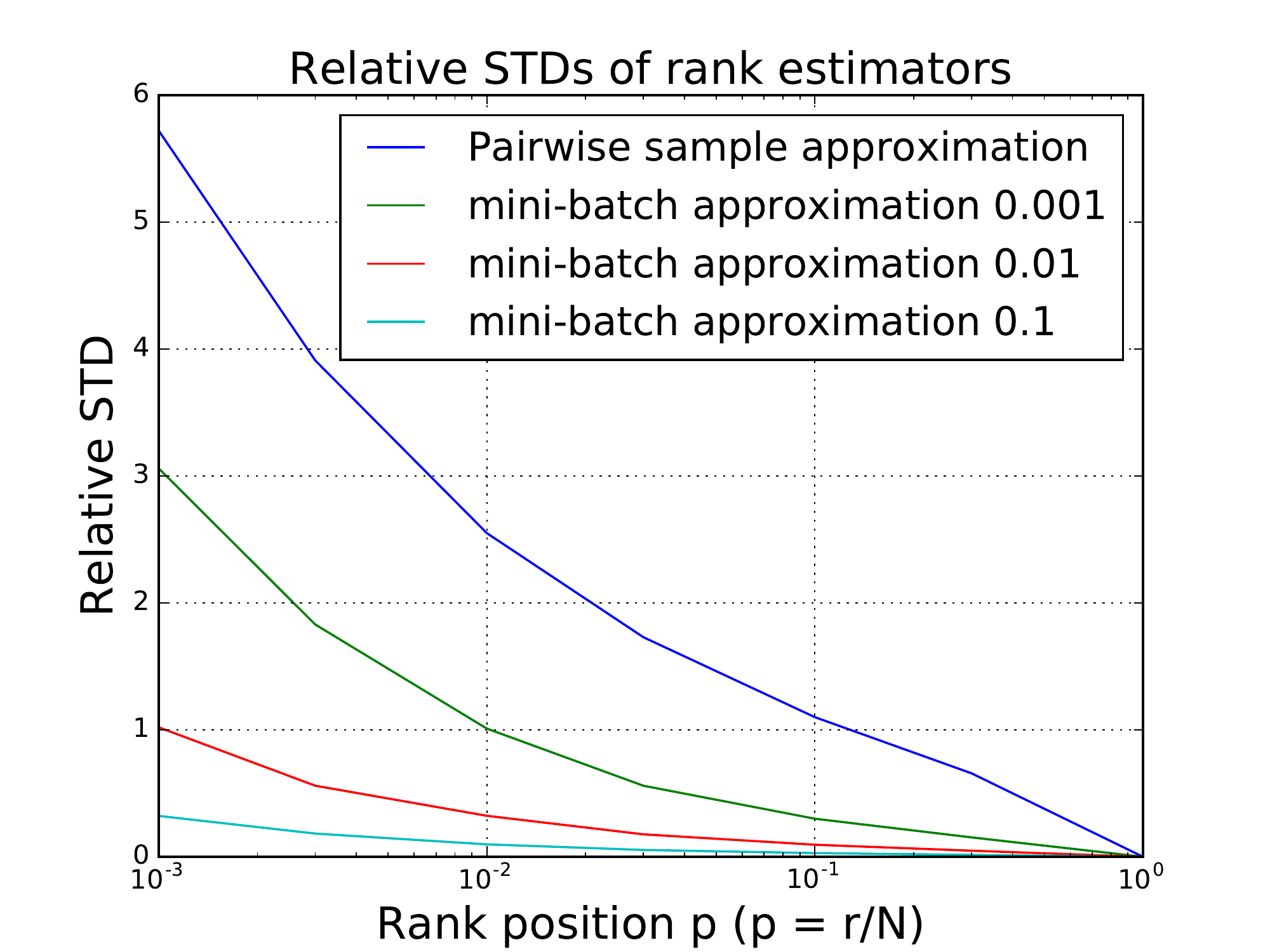}
\caption{Standard deviations (relative values) of two types of rank estimators at different item ranks. Simulation is done with item set size N=100,000. `online' uses estimator (\ref{eq:warp3}) and `sampled-batch q' uses (\ref{eq:mrmini}) where $q=|\textbf{Z}|/|\textbf{Y}|$.}
\label{pic:std}
\end{figure}

Although a sampling step is involved, we argue below that (\ref{eq:mrmini}) does not lead to large variances as the sampling in pairwise approaches do. First, (\ref{eq:mrmini}) is an unbiased estimator of (suppressed) margin rank. Second, the sampling schemes in the two approaches are different. To have a better idea, we conduct simulation of two types of sampling and plot standard deviations in Figure \ref{pic:std} shows that (\ref{eq:mrmini}) has much smaller variances than (\ref{eq:warp3}) as long as $|\textbf{Z}|/|\textbf{Y}|$ is not too small.

\subsection{Smooth rank-sensitive loss functions}

Rank-based loss function (\ref{eq:owc}) operates on item ranks and provides a mechanism to advocate top accuracy. However, its input has to be an integer. It is also non-smooth. Consequently, it is not applicable to our batch rank estimators and does not support gradient-based optimization. In the following, we generalize (\ref{eq:owc}) to smooth rank-sensitive loss functions that similarly advocate top accuracy. 

We first observe that a loss function $\ell$ would encourage top accuracy when the following conditions are satisfied:
\begin{itemize}
\item $\ell$ is a smooth function of the rank r -- rank sensitive (rs).
\item $\ell$ is increasing, i.e., $\ell' > 0$.
\item $\ell$ is concave, i.e., $\ell'' < 0$.
\end{itemize}

The second condition indicates the loss increases when the rank of a target item increases. Thus, given a single item, minimizing the objective would try to push the item to the top. The third condition means the increase is fast when rank is small and slow when the rank is large. So it is more sensitive to small ranking items. Given more than one item, an algorithm that minimizes this objective would prioritize on those items with small estimated rank values.

Based on the observation, we study several types of functions $\ell^{rs}$ that satisfy these conditions: polynomial functions, logarithmic functions, and exponential functions, i.e., 
\begin{itemize}
\item $\ell_1^{rs}(r) = (1+r)^p, \quad 0<p<1$
\item $\ell_2^{rs}(r) = \log(r+1)$ 
\item $\ell_3^{rs}(r) = 1 - \lambda^{-r}, \quad \lambda > 1$
\end{itemize}

It follows standard calculus to verify that these functions satisfy the above conditions. Thus they all incur (smoothly) weighed loss based on estimated rank values and advocate top accuracy. We plot part of these functions in Figure \ref{fig:rsl} and compare them to BPR and OWA (with ${\alpha}s=1, 1/2, 1/3, ...$). BPR is equivalent to a linear function which places a uniformly increasing penalty on the estimated rank. Polynomial and logarithmic functions have a diminishing return when the estimated rank increases and is unbounded. Exponential functions are bounded by 1, and the penalty on high rank values is quickly saturated.

\subsection{Algorithm}
An algorithm can be then formulated as minimizing an objective based on the (mini-)batch approximated rank values and rank sensitive loss functions. It sums over all pairs of observed user-item activity. Particularly, 
\begin{equation}
\label{eq:alg}
L = \sum_{(x,y)\in \textbf{S}} \ell^{rs}(r^{mb}_y(f,x,\textbf{y})) + \phi(f),
\end{equation}
where $r_y^{mb}$ is given by (\ref{eq:mrmini}) and $\ell^rs$ takes $\ell^{rs}_i, i=1,2,3$. $\phi(f)$ is a model regularization term.

Gradient based methods are used to solve the optimization problem. The gradient with respect to the model can be written as (ignore the regularization term for the moment)
\begin{equation}
\label{eq:gradient}
\frac{\partial L}{\partial f} = \sum \ell'(r) \times \frac{\partial r}{\partial f},
\end{equation}
where $\ell'(r)$ takes the form $p(1+r)^{p-1}$ (or $1/(r+1)$, or $\lambda^{-r}$). We call the framework defined in (\ref{eq:alg}) Batch-Approximated-Rank-Sensitive loss (BARS). The detailed algorithm is described in Algorithm \ref{alg}.

\begin{algorithm}[!ht]
\SetAlgoLined
\KwIn{Training data $\textbf{S}$; mini-batch size $m$; Sample rate $q$; a learning rate $\eta$.}
\KwOut{The model parameters f.}
 initialize parameters of model f randomly\;
 \While{Objective (\ref{eq:alg}) is not converged}{
  sample a mini-batch of observations $\{(x,y)_i\}_{i=1}^m$\;
  sample item subset $\textbf{Z}$ from $\textbf{Y}$, $q=|\textbf{Z}|/|\textbf{Y}|$\;
  compute approximated ranks by (\ref{eq:mrmini})\;
  update model f parameters: \\
  $\qquad f = f - \eta * \partial \ell / \partial f$ based on (\ref{eq:gradient})\;
 }
 \caption{}
 \label{alg}
\end{algorithm}

Note that in Algorithm \ref{alg} computation is conducted in a batch manner. Particularly, it computes model scores between a mini-batch of users ($\textbf{x}$) and sampled items ($\textbf{Z}$) in parallel. Every step it updates parameters of all the users in $\textbf{x}$ and items in $\textbf{Z}$.

\textit{Comparisons to Lambda-based methods.} Lambda-based methods such as LambdaNet and LambdaRank use an updating strategy in the following form
\begin{equation}
\label{eq:lambda}
\delta f_{ij} = \lambda_{ij} * |\Delta NDCG_{ij}|,
\end{equation} 
where $\lambda_{ij}$ is a regular gradient term and $|\Delta NDCG_{ij}|$ is the NDCG difference of the ranking list for a specific user if the positions (ranks) of items i, j get switched and acts as a  scaling term that is motivated to directly optimize the ranking measure.

Compare (\ref{eq:gradient}) and (\ref{eq:lambda}), $\ell'(r)$ replaces $|\Delta NDCG_{ij}|$ and functions in a similar role. In our cases, $\ell'(r)$ is decreasing in r. Thus it gives a higher incentive to fix errors in low-ranking items. However, instead of directly computing the NDCG difference which can be computationally expensive in large scale settings, the proposed algorithm first approximates the rank and then computes the scaling value through derivative of the loss function.

\section{Related work}
\label{sec:related}

\textbf{Top accuracy in traditional ranking.} Top accuracy is a classical challenge for ranking problems. 
One typical type of approaches is to develop smooth surrogates of the non-smooth ranking metric.\cite{weimer2008cofi} use structure learning and propose smooth convex upper bounds of a ranking metric. \cite{taylor2008softrank} develops a smooth approximation based on connections between score distribution and rank distribution. Similarly, in a kernel SVM setting, \cite{agarwal2011infinite} proposes a formulation based on infinity norm. \cite{boyd2012accuracy} generalizes the notion of top k to top $\tau$-quantile and optimizes a convex surrogate of the corresponding ranking loss.

Alternatively, \cite{burges2005learning} starts from the average precision loss and modifies model updating steps to promote top accuracy. Similar ideas are developed in \cite{burges2007learning,wu2010adapting}. \cite{burges2007learning} writes down the gradient directly rather than derives from a loss. \cite{wu2010adapting} works on a boosted tree formulation.

\noindent\textbf{Personalized ranking.} Early works on personalized ranking do not necessarily focus on top accuracy. \cite{hu2008collaborative} first studies the task and converts it as a regression problem. \cite{rendle2009bpr} introduces ranking based optimization and optimizes a criterion similar to AUC. \cite{rendle2014improving} work on improving the efficiency but do not change the learning criterion.

To promote top accuracy and deal with large scale settings, \cite{shi2012tfmap} develops rank approximation based on model score and proposes a smooth approximation of MAP. \cite{shi2012climf} adopts the same idea and targets at MRR. 

Pairwise algorithms~\cite{weston2010large,yuan2016lambdafm}~are then proposed to estimate ranks through sampling methods. \cite{weston2010large} updates the model based on an operator Ordered Weighted Average. \cite{yuan2016lambdafm} uses a similar idea as in \cite{burges2005learning}.

\section{Experiments}
\label{sec:exp}

In this section we conduct experiments on three large scale real-world datasets to verify the effectiveness of the proposed methods.

\subsection{Experimental setup}
\subsubsection{Dataset}

We validate our approach on three public datasets from different domains: 1) movie recommendation. 2) business reviews at Yelp; 3) job recommendation from XING\footnote{www.xing.com}; We describe the datasets in detail below.

\noindent\textbf{MovieLens-20m} The dataset has anonymous ratings made by MovieLens users.\footnote{www.movielens.org} We transform the data into binary indicating whether a user rated a movie above $4.0$. We discard users with less than 10 movie ratings and use 70\%/30\% train/test splitting. Attributes include movie genres and movie title text.

\noindent\textbf{Yelp} dataset comes from Yelp Challenge.\footnote{https://www.yelp.com/dataset\_challenge. Downloaded in Feb 17.} We work on recommendation related to which business a user might want to review. Following the \textit{online protocol} in \cite{he2016fast}, we sort all interactions in chronological order and take the last 10\% for testing and the rest for training. Business items have attributes including \textit{city}, \textit{state}, \textit{categories}, \textit{hours}, and \textit{attributes} (e.g. ``valet parking,'' ``good for kids'').

\noindent\textbf{XING} contains about 12 weeks of interaction data between users and items on XING. Train/test splitting follows the RecSys Challenge 2016 \cite{abel2016recsys} setup where the last two weeks of interactions for a set of 150,000 \textit{target users} are used as test data. Rich attributes are associated with the data like career levels, disciplines, locations, job descriptions etc. Our task is to recommend to users a list of job posts that they are likely to interact with.

We report dataset statistics in Table~\ref{t:data1}.

\begin{table}
\centering
\small{
\begin{tabular}{|c|c|c|c|c|}
\hline
\textbf{Data}   & $\left|U\right|$ & $\left|I\right|$ & $\left|S_{train}\right|$ & $\left|S_{test}\right|$ \\
\hline
\movietwenty   & 138,493  & 27,278 & 7,899,901 & 2,039,972  \\ \hline
\yelp       & 1,029,433     & 144,073  &  1,917,218 & 213,460  \\ \hline
\xing       & 1,500,000     & 327,002  & 2,338,766  & 484,237  \\ \hline
\end{tabular}
}
\caption{Dataset statistics. U: users; I: items; S: interactions.}
\label{t:data1}
\end{table}

\begin{table*}[!t]
\centering
\begin{tabular}{|c||c|c|c||c|c|c||  c| c| c|} \hline
\textbf{Datasets}  & \multicolumn{3}{c||}{\movietwenty} & \multicolumn{3}{|c||}{\yelp}  & \multicolumn{3}{|c|}{\xing} \\ \hline
\textbf{Metrics}  & \pre  & \rec  & \ndcg  & \pre  & \rec    & \ndcg  & \pre  & \rec  & \ndcg  \\ \hline
 \pop         &6.2&10.0&8.5  &0.3&0.9&0.5    &0.5&2.7&1.3  \\ \hline
\bpr          &6.1 &10.2 & 8.3 &0.1 & 0.4&0.2 &0.3 &2.2 &0.9 \\ \hline
\bbpr     &9.3 &14.3 &12.9  &0.9&3.4&1.9 &1.3& 9.2 & 4.2  \\ \hline
\awarp      &\textit{10.1}&13.3&13.5 &1.3&4.3&2.5     &2.6&11.6 &6.7\\ \hline
CE      &9.6&14.3&13.2  &\textit{1.4}&4.5&2.6  &2.5& \textit{12.3} & 6.5 \\ \hline\hline
SR-log        &9.9 & 14.5  &\textit{13.6}  &\textit{1.4} & \textit{5.2} &\textit{2.9}  &\textit{2.8}  &\textit{12.3}  &\textit{6.9} \\ \hline
MR-poly      &\textbf{10.2} &\textbf{14.8} &\textbf{13.9}    &\textbf{1.5} &\textit{5.2} & \textit{2.9}  &\textit{2.8}  &\textbf{12.5}  &\textit{6.9}\\ \hline
MR-log        &\textbf{10.2}&\textit{14.6}&\textbf{13.9}  &\textbf{1.5}& 5.1& \textit{2.9}   &\textbf{2.9}& \textbf{12.5}&\textbf{7.1}\\ \hline
SMR-log      &\textbf{10.2} &\textit{14.6} &\textbf{13.9}   &\textbf{1.5}& \textbf{5.4}& \textbf{3.0}  &\textbf{2.9}& \textbf{12.5} & \textbf{7.1} \\ \hline
\end{tabular}
\caption{Recommendation accuracy comparisons (in \%). Results are averaged over 5 experiments with different random seeds. Best and second best numbers are in bold and italic, respectively.}
\label{t:accuracy}
\end{table*}

\subsubsection{Methods}
We study multiple algorithms under our learning framework and compare them to various baseline methods. Particularly, we study the following algorithms:

\begin{itemize}
\item \pop.~A naive baseline model that recommends items in terms of their popularity. 
\item \bpr\cite{rendle2009bpr}. BPR optimizes AUC and is a widely used baseline.
\item \bbpr. A batch version of BPR. It uses the same logistic loss but updates a target item and a batch of negative items every step.
\item \awarp\cite{weston2010large,hong2013co}. A state-of-the-art pairwise personalized ranking method.
\item CE. Cross entropy loss is recently used for item recommendation in \cite{hidasi2015session,covington2016deep}. 
\item SR-log. The proposed algorithm with the smoothed rank approximation without margin formulation (sr) and logarithmic function (log).
\item MR-poly. The proposed algorithm with the margin rank approximation (mr) and polynomial function (poly).
\item MR-log. The proposed algorithm with the margin rank approximation (mr) and logarithmic function (log).
\item SMR-log. The proposed algorithm with the suppressed margin rank approximation (smr) and logarithmic function (log).
\end{itemize}

\bpr~and~\awarp~are implemented by~\lightfm \cite{kula_metadata_2015}). We implemented the other algorithms. 
 
We apply the algorithms to hybrid matrix factorization~\cite{shmueli2012care}, a factorization model that represents users and items as linear combinations of their attribute embedding. Therefore model parameters $f$ include factors of users, items, and their attributes.

Early stopping is used on a development dataset split from training for all models. Hyper-parameter model size is tuned in \{10, 16, 32, 48, 64\}; learning rate is tuned in \{0.5, 1, 5, 10\}; when applicable, dropout rate is 0.5. Batch based approaches are implemented based on Tensorflow 1.2 and run on a single GPU (NVIDIA Tesla P100 GPU ).~\lightfm~runs on Cython with 5 cores CPU (Intel Xeon 3.30GHz).

\subsubsection{Metrics}
We assess the quality of recommendation results by comparing models' recommendation to ground truth interactions and report \textit{Precision} (P), \textit{Recall} (R) and \textit{Normalized Discounted Cumulative Gain} (NDCG). We report scores \textit{after removing historical items from each user's recommendation list} on~\yelp~and \movietwenty~datasets because users seldom re-interact with items in these scenarios (Yelp reviews/movie ratings). This improves performance for all models but does not change relative comparison results.

\subsection{Results}

\subsubsection{Quality of rank approximations}
\begin{figure}
\centering
\begin{subfigure}{.25\textwidth}
  \centering
  \includegraphics[width=.99\linewidth]{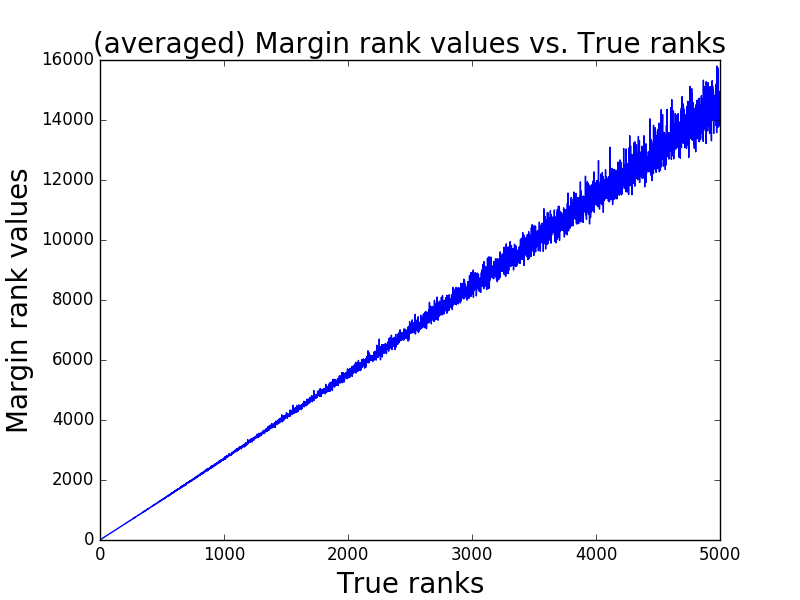}
  \caption{0-50000.}
  \label{fig:qual1}
\end{subfigure}%
\begin{subfigure}{.25\textwidth}
  \centering
  \includegraphics[width=.99\linewidth]{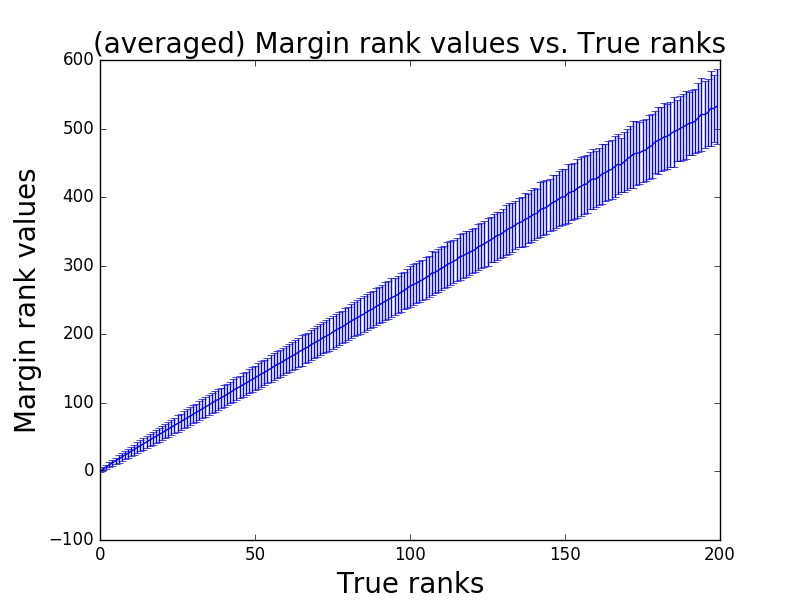}
  \caption{0-200.}
  \label{fig:qual2}
\end{subfigure}
\caption{Approximated rank values compared to the true rank values. \ref{fig:qual2} is a zoomed-in version with error bars.}
\label{fig:rankapprox}
\end{figure}

We first study how well the proposed methods approximate the true item ranks. To do that, we run one epoch of training on~\xing~dataset and compute the values in (\ref{eq:mr}) and the true item rank. We plot the value of (\ref{eq:mr}) as a function of the true rank in Figure \ref{fig:rankapprox}.

Figure \ref{fig:qual1} shows in a very large range (0-50000) the estimator in (\ref{eq:mr}) is linearly aligned with the true item rank, especially when true item ranks are small -- note those regions are what we care most about. We further zoom into the top region (0-200) and plot error bars in addition to function mean values. In Figure \ref{fig:qual2}, we see limited variances. For example, the relative standard deviation is smaller than 0.1, which indicates stable rank approximation. As a comparison, the simulation in Figure \ref{pic:std} suggests stochastic pairwise estimation should lead to relative standard deviation much more than 6.

\subsubsection{Recommendation accuracy}

Recommendation accuracy is reported in Table \ref{t:accuracy}. We have the following observations.

First, vanilla pairwise~\bpr~performs poorly due to large itemset size. In contrast, batch version~\bbpr~bypasses the difficulty of sampling negative items and updates model parameters smoothly, achieving decent accuracies.

Second,~\awarp~and CE outperforms~\bbpr. Both methods target more than averaged precision. Compared with each other, CE penalizes more on a correct item ranking behind, thus showing consistent better performances in recall.

Third, the proposed methods consistently outperform~\awarp~and CE. Compared to CE, the improvements suggest the effectiveness of the rank sensitive loss, which fits better than the classification loss. Compared to~\awarp, we attribute the improvement to better rank approximations and possibly other factors like smoother parameter updates.

Finally, we compare the different variants of the proposed methods. SR-log underperforms and it suggests the benefit of margin formulation. Polynomial loss functions have similar results compared to logarithmic functions but require a bit tuning in the hyper-parameter $p$. Suppress margin rank (SMR) performs a bit better than MR probably due to its better rank approximation.

\subsubsection{Time efficiency}

We study the time efficiency of the proposed methods and compare to pairwise algorithm system. Note that we are \textit{not} just interested in the comparisons of the absolute numbers. Rather, we focus on the \textbf{trend} how time efficiency changes when data scale increases.

We characterize the dataset complexity in Table \ref{t:time} by the density (computed by $\frac{\#. observations}{\#. users \times \#. items}$), the number of total parameters, the average number of attributes per user-item pair, and the itemset size. From Table \ref{t:time},~\movietwenty~has the densest observations, the smallest number of total parameters and attributes per user-item, and the smallest itemset size. Thus we call it ``small'' in complexity. Conversely, we call~\xing~``large'' in complexity. \yelp~is between the two and called ``medium''.

\begin{figure}
\centering
\begin{subfigure}{.25\textwidth}
  \centering
  \includegraphics[width=.99\linewidth]{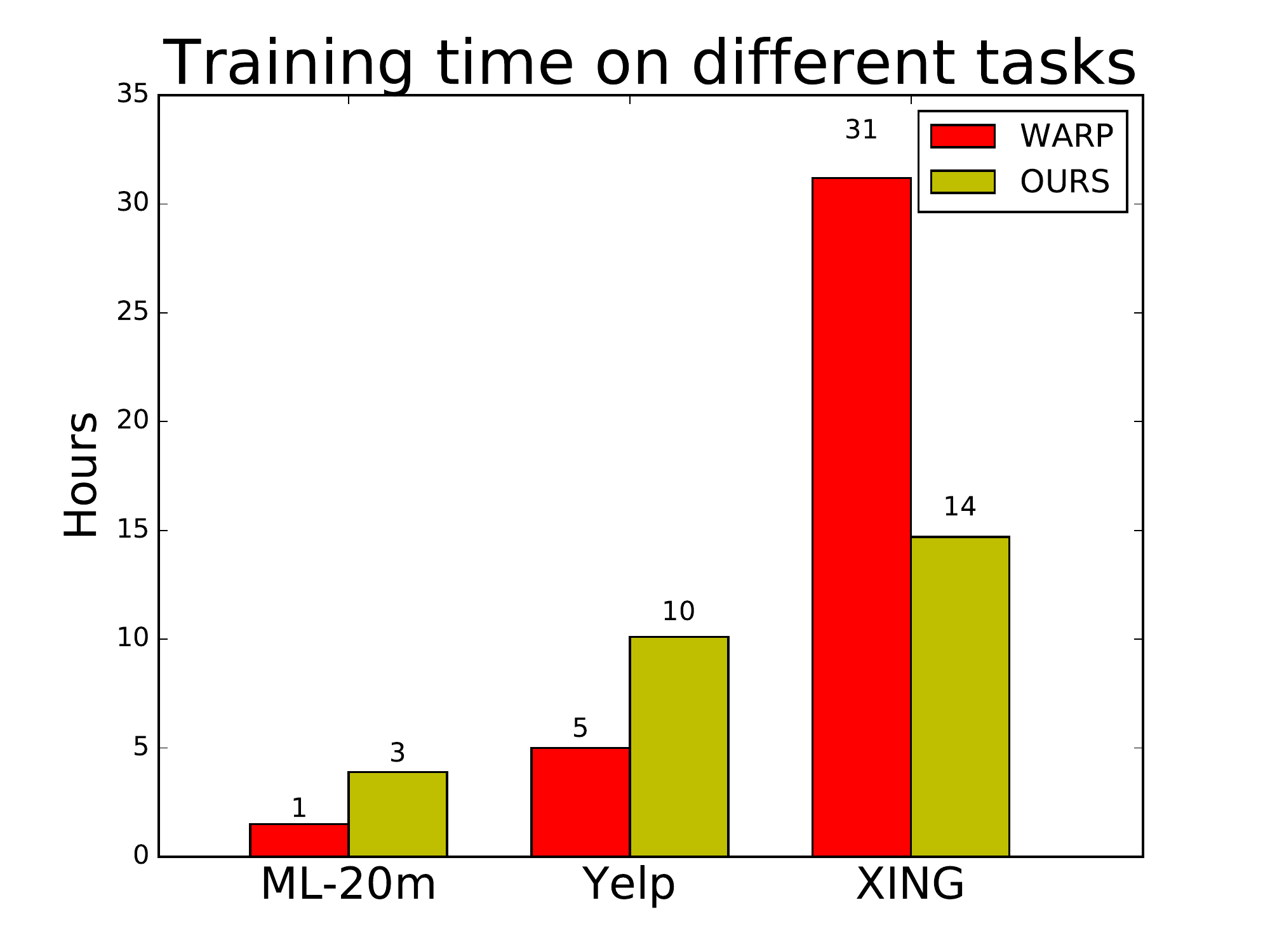}
  \caption{Across datasets.}
  \label{fig:tconv}
\end{subfigure}%
\begin{subfigure}{.25\textwidth}
  \centering
  \includegraphics[width=.99\linewidth]{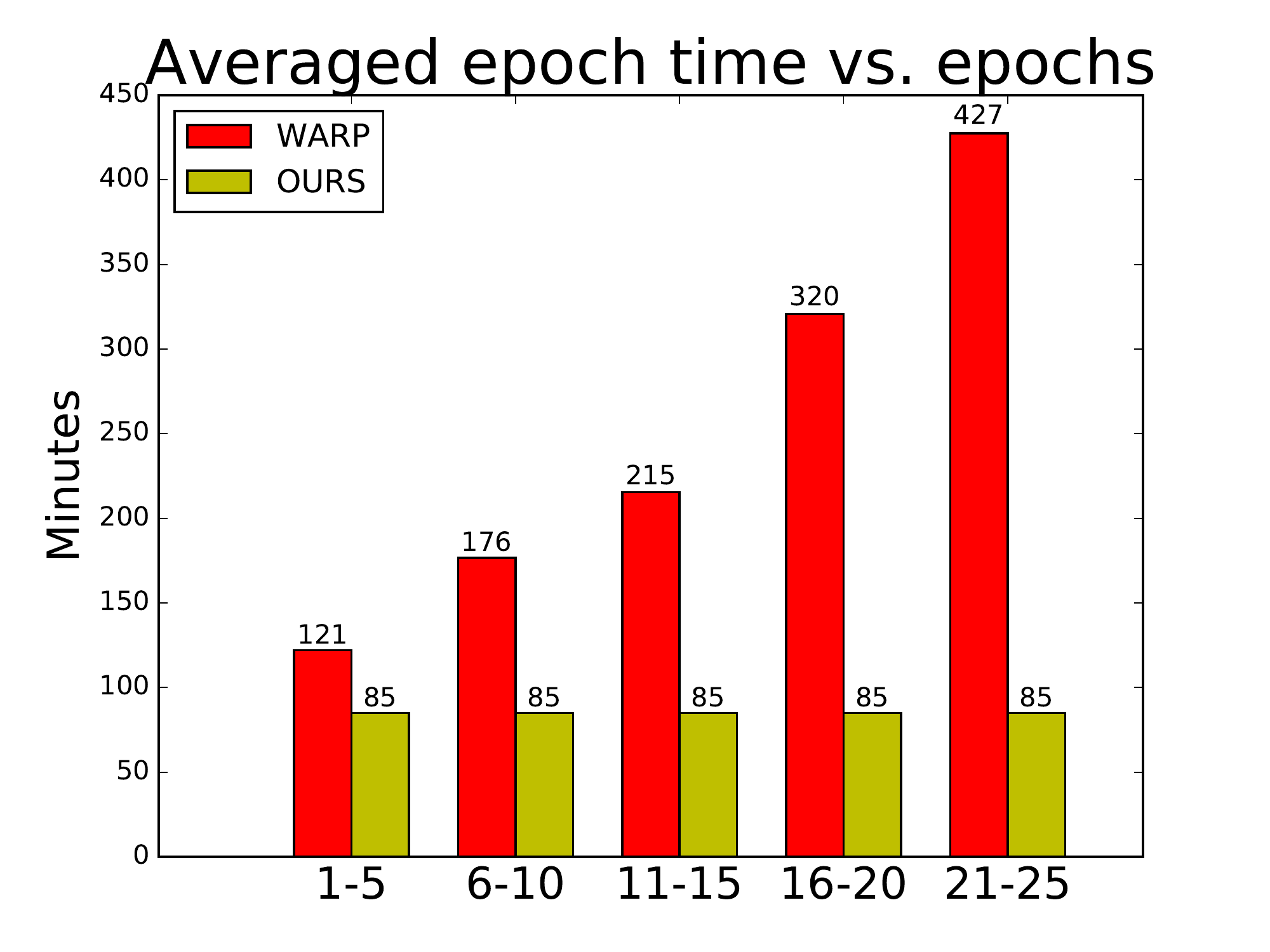}
  \caption{Across time.}
  \label{fig:tepoch}
\end{subfigure}
\caption{Training time comparisons between WARP and BARS. Fig. \ref{fig:tconv} plots how training time changes across datasets with different scales; Fig. \ref{fig:tepoch} plots how epoch time changes as the training progresses. }
\label{fig:time}
\end{figure}

Two results are reported in Figure \ref{fig:time}. Figure \ref{fig:tconv} shows across different datasets the converging time needed to reach the best models from the two systems: WARP and BARS (ours). WARP takes a shorter time in both ``small'' and ``medium'' but its running time increases very fast; BARS increases slowly in the training time and wins over at ``large''. 

Figure \ref{fig:tepoch} depicts the averaged epoch training time of the two models. BARS has a constant epoch time. In contrast, WARP keeps increasing the training time. This is expected because when the model becomes better, it takes a lot more sampling iterations every step.

\begin{table}[t]
\centering
\begin{tabular}{|c||c|c|c|} \hline
\textbf{Datasets} & \movietwenty & \yelp & \xing \\ \hline
 Density & 2.6e-3 & 1.4e-5 & 5.8e-6 \\ \hline
 \# of param. & 4.6M & 9.3M & 12.1M \\ \hline
 \# of Attr.      & 11     & 19     & 33 \\ \hline
$\left| I \right|$  & 27K &144K & 327K \\ \hline \hline
complexity    & small & medium & large \\ \hline
\end{tabular}
\caption{Dataset/model complexity comparisons.}
\label{t:time}
\end{table}

\subsubsection{Robustness to mini-batch size}

The full batch algorithm is used in the above experiments. We are also interested to see how it performs with a sampled batch loss. In Table \ref{t:sampled}, we report loss values and~\ndcg~scores on development split and compare them to full batch versions. With the sampling proportion 0.1 or 0.05, sampled version algorithm gives almost identical results as the full batch version on all datasets. This suggests the robustness of the algorithm to mini-batch size.

\begin{table}
\centering
\begin{tabular} { |c|| c | c | c | c| c|c|} \hline
 \multirow{ 2}{*}{q}	         & \multicolumn{2}{|c|}{\movietwenty}  & \multicolumn{2}{|c|}{\yelp} & \multicolumn{2}{|c|}{\xing}  \\ \cline{2-7}
                  &  obj & \small{NDCG} & obj & \small{NDCG} & obj & \small{NDCG} \\ \hline
1.0    & 6.49 & 16.0  & 7.94 & 3.0   &  6.42  & 9.9\\ \hline
0.1  & 6.47 & 15.9    & 7.87  & 3.0  & 6.40  & 10.0 \\ \hline
0.05    & 6.47 & 15.9    & 7.90   &2.9  & 6.42  & 9.8\\ \hline
\end{tabular}
\caption{Comparisons of objective values (obj) and recommendation accuracies (NDCG) on development set among full batch and sampled batch algorithms. $q=|\textbf{Z}|/|\textbf{Y}|$, $q=1.0$ means full batch.} \label{t:sampled}
\end{table}

\section{Conclusion}

In this work we address personalized ranking task and propose a learning framework that exploits the ideas of batch training and rank-dependent loss functions. The proposed methods allow more accurate rank approximations and empirically give competitive results.

In designing the framework, we have purposely tailored our approach to using parallel computation and support back-propagation updates. This readily lends itself to flexible models such as deep feedforward networks, recurrent neural nets, etc. In the future, we are interested in exploring the algorithm in the training of deep neural networks.

\label{sec:summary}

\clearpage
\bibliographystyle{aaai}
\bibliography{main}

\end{document}